\documentclass[10pt,twocolumn,letterpaper]{article}

\usepackage[article]{cvpr}
\usepackage{tikz}
\usetikzlibrary{shapes.geometric, arrows.meta, positioning}
\usepackage{graphicx}
\usepackage{amsmath}
\usepackage{amssymb}
\usepackage{booktabs}
\usepackage{ctable}
\usepackage{multirow}
\usepackage{array}
\usepackage{colortbl}
\usepackage{makecell}
\usepackage[ruled,vlined,linesnumbered]{algorithm2e}
\usepackage{subcaption}

\definecolor{cvprblue}{rgb}{0.21,0.49,0.74}
\usepackage[pagebackref,breaklinks,colorlinks,citecolor=cvprblue]{hyperref}
\graphicspath{{img/}{../img/}{./}{../}}

\def\paperID{6}
\def\confName{Interactive Physical AI}
\def\confYear{}

\begin{document}
	
	\title{Resource-Efficient Iterative LLM-Based NAS with Feedback Memory}
	
	\author{Xiaojie Gu,\space\space\space Dmitry Ignatov,\space\space\space Radu Timofte\\
		\small{Computer Vision Lab, CAIDAS \& IFI, University of W\"urzburg, Germany}}
	\maketitle

	%%%%%%%%% ABSTRACT
	\begin{abstract}
    Neural Architecture Search (NAS) automates network design, but conventional methods demand substantial computational resources. We propose a closed-loop pipeline leveraging large language models (LLMs) to iteratively generate, evaluate, and refine convolutional neural network architectures for image classification on a single consumer-grade GPU without LLM fine-tuning. Central to our approach is a historical feedback memory inspired by Markov chains: a sliding window of $K{=}5$ recent improvement attempts keeps context size constant while providing sufficient signal for iterative learning. Unlike prior LLM optimizers that discard failure trajectories, each history entry is a structured diagnostic triple---recording the identified problem, suggested modification, and resulting outcome---treating code execution failures as first-class learning signals. A dual-LLM specialization reduces per-call cognitive load: a \emph{Code Generator} produces executable PyTorch architectures while a \emph{Prompt Improver} handles diagnostic reasoning. Since both the LLM and architecture training share limited VRAM, the search implicitly favors compact, hardware-efficient models suited to edge deployment. We evaluate three frozen instruction-tuned LLMs (${\leq}7$B parameters) across up to 2000 iterations in an unconstrained open code space, using one-epoch proxy accuracy on CIFAR-10, CIFAR-100, and ImageNette as a fast ranking signal. On CIFAR-10, DeepSeek-Coder-6.7B improves from 28.2\% to 69.2\%, Qwen2.5-7B from 50.0\% to 71.5\%, and GLM-5 from 43.2\% to 62.0\%. A full 2000-iteration search completes in ${\approx}18$ GPU hours on a single RTX~4090, establishing a low-budget, reproducible, and hardware-aware paradigm for LLM-driven NAS without cloud infrastructure.
	\end{abstract}
	
	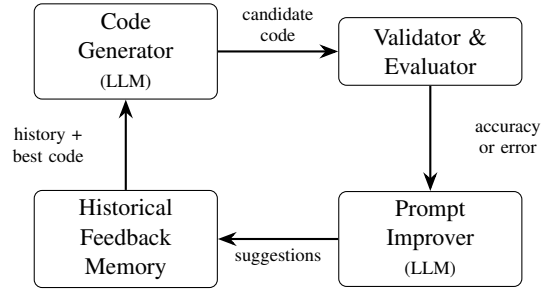
\begin{figure}[t]
		\centering
		\begin{tikzpicture}[
			node distance=1.4cm and 1.6cm,
			block/.style={rectangle, draw, rounded corners=3pt, text width=2.2cm, text centered, minimum height=0.9cm, font=\small},
			arrow/.style={-{Stealth[length=2.5mm]}, thick},
			label/.style={font=\scriptsize, text width=1.8cm, align=center}
			]
			\node[block] (gen) {Code\\Generator\\{\scriptsize(LLM)}};
			\node[block, right=of gen] (eval) {Validator \&\\Evaluator};
			\node[block, below=of eval] (imp) {Prompt\\Improver\\{\scriptsize(LLM)}};
			\node[block] at (gen |- imp) (hist) {Historical\\Feedback\\Memory};
			
			\draw[arrow] (gen) -- node[above, label] {candidate\\code} (eval);
			\draw[arrow] (eval) -- node[right, label] {accuracy\\or error} (imp);
			\draw[arrow] (imp) -- node[below, label] {suggestions} (hist);
			\draw[arrow] (hist) -- node[left, label] {history +\\best code} (gen);
		\end{tikzpicture}
		\caption{Overview of the iterative NAS pipeline. The Code Generator produces a candidate architecture as executable PyTorch code. The Evaluator validates and trains it using one-epoch proxy evaluation. The Prompt Improver analyzes results with historical feedback memory to generate targeted improvement suggestions for the next iteration.}
		\label{fig:pipeline}
	\end{figure}
	
	%%%%%%%%% BODY TEXT
	\section{Introduction}
	\label{sec:intro}

	\begin{figure*}[t!]
		\centering
		\begin{subfigure}[b]{0.325\textwidth}
			\centering
			\includegraphics[width=\linewidth]{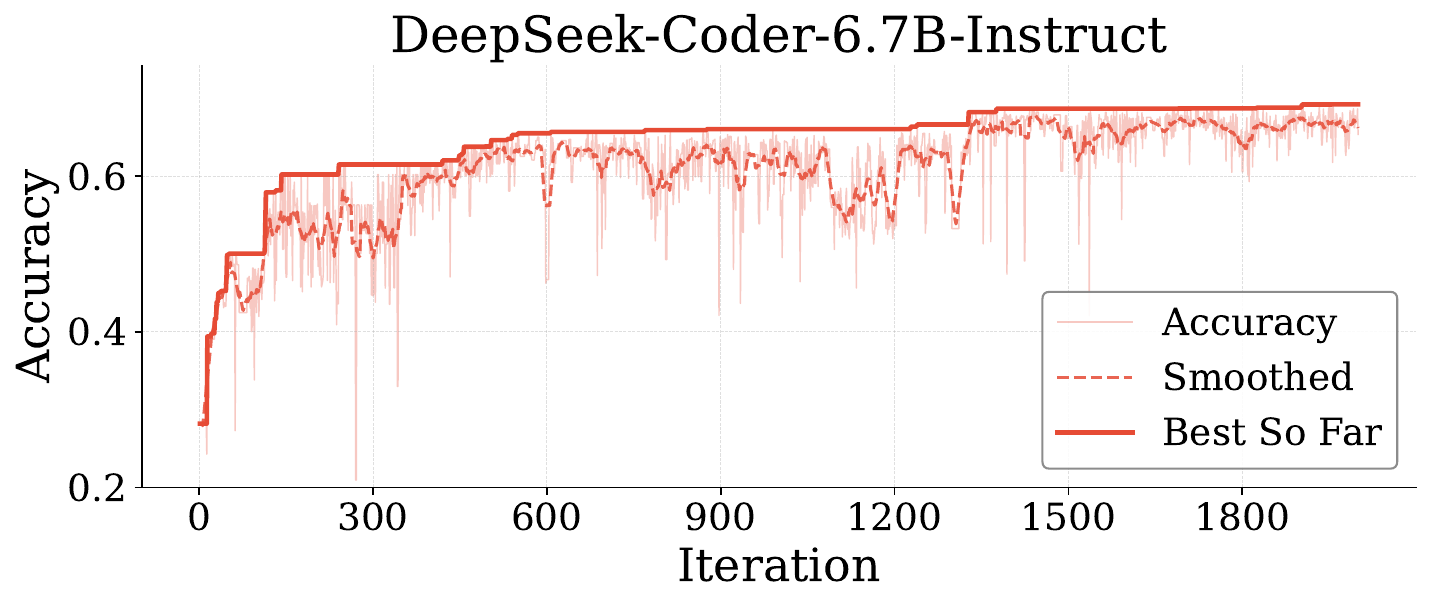}
			\caption{DeepSeek-Coder (CIFAR-10)}
			\label{fig:acc_deepseek_cifar}
		\end{subfigure}
		\hfill
		\begin{subfigure}[b]{0.325\textwidth}
			\centering
			\includegraphics[width=\linewidth]{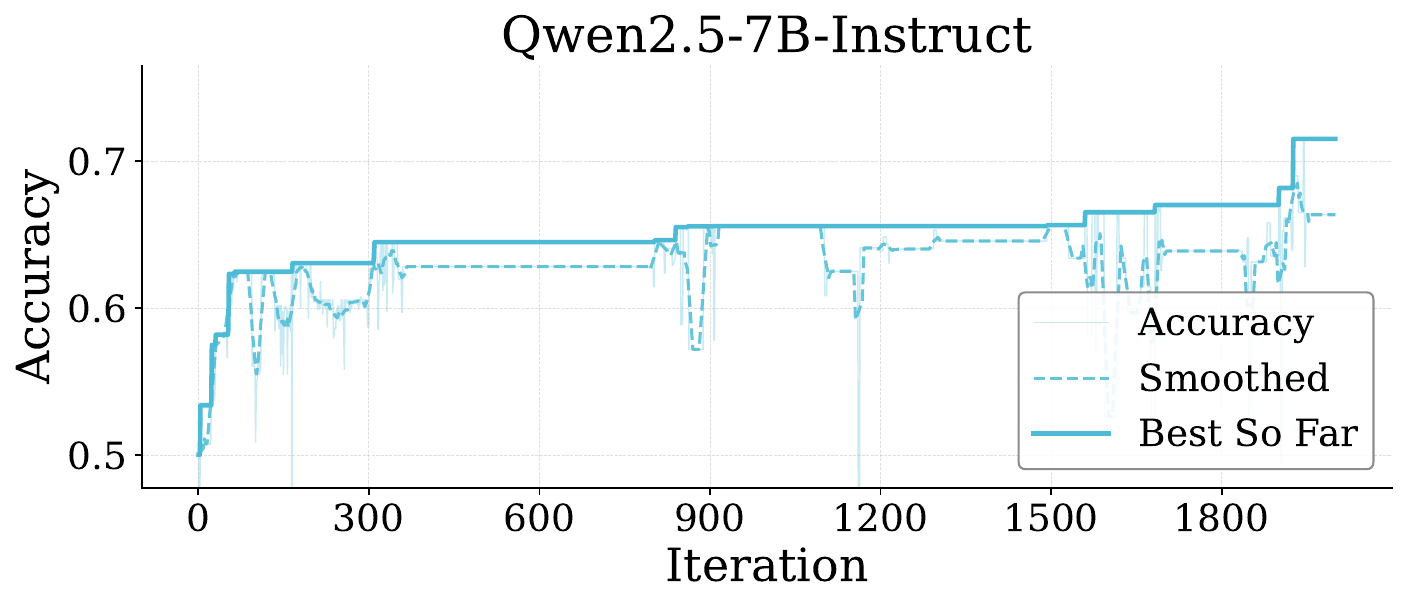}
			\caption{Qwen2.5 (CIFAR-10)}
			\label{fig:acc_qwen_cifar}
		\end{subfigure}
		\hfill
		\begin{subfigure}[b]{0.325\textwidth}
			\centering
			\includegraphics[width=\linewidth]{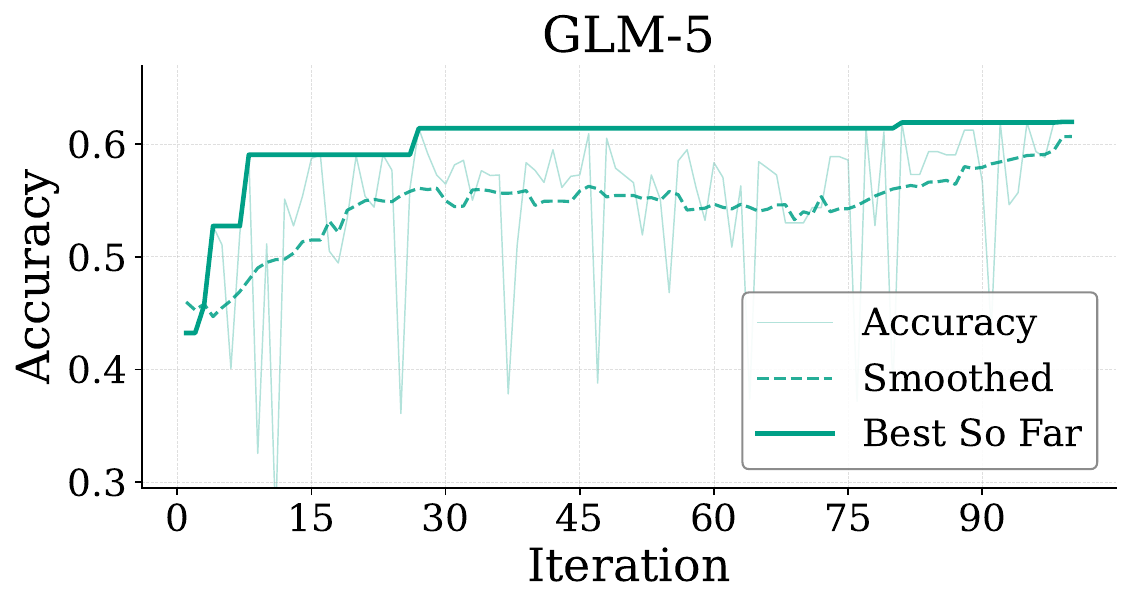}
			\caption{GLM-5 (CIFAR-10)}
			\label{fig:acc_glm5_cifar}
		\end{subfigure}

		\vspace{0.3cm}
		
		\begin{subfigure}[b]{0.325\textwidth}
			\centering
			\includegraphics[width=\linewidth]{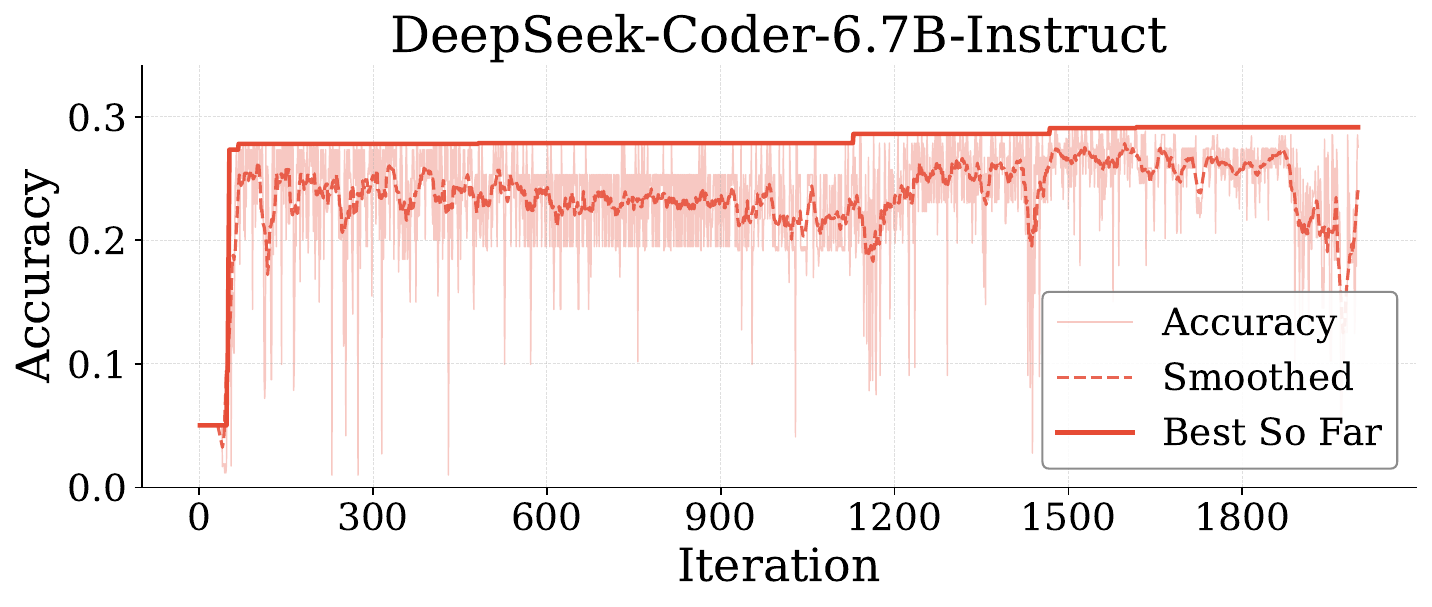}
			\caption{DeepSeek-Coder (CIFAR-100)}
			\label{fig:acc_deepseek_cifar100}
		\end{subfigure}
		\hfill
		\begin{subfigure}[b]{0.325\textwidth}
			\centering
			\includegraphics[width=\linewidth]{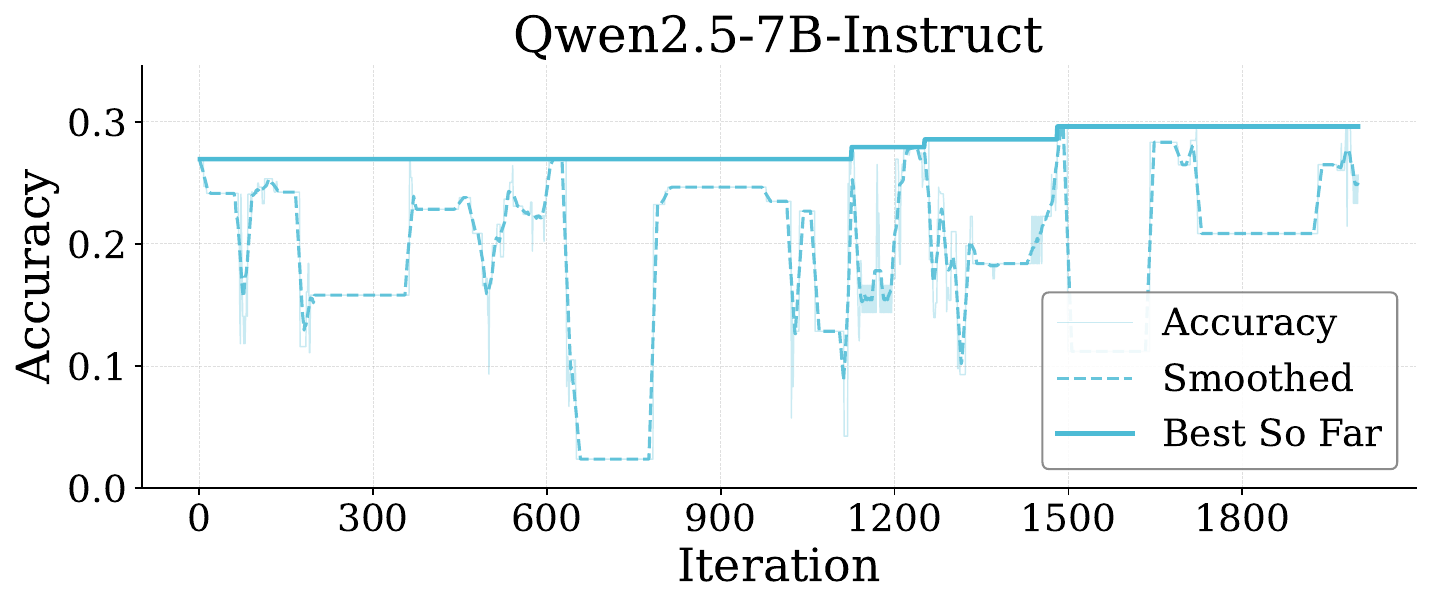}
			\caption{Qwen2.5 (CIFAR-100)}
			\label{fig:acc_qwen_cifar100}
		\end{subfigure}
		\hfill
		\begin{subfigure}[b]{0.325\textwidth}
			\centering
			\includegraphics[width=\linewidth]{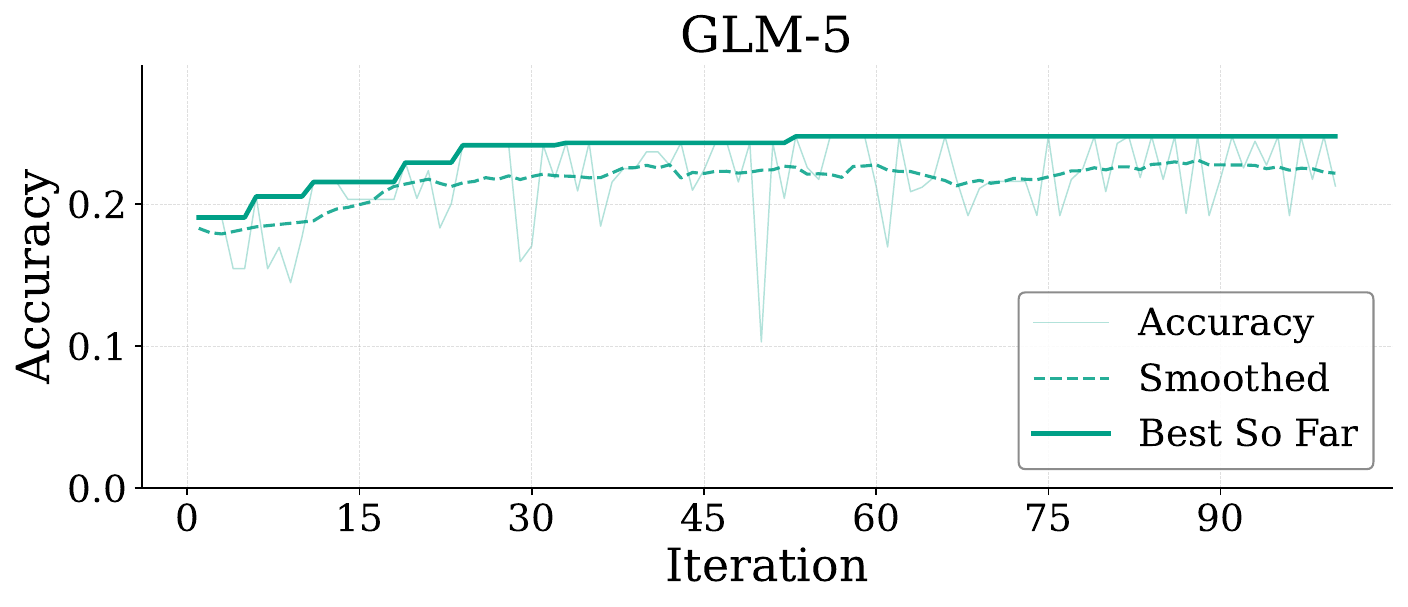}
			\caption{GLM-5 (CIFAR-100)}
			\label{fig:acc_glm5_cifar100}
		\end{subfigure}

		\vspace{0.3cm}
		
		\begin{subfigure}[b]{0.325\textwidth}
			\centering
			\includegraphics[width=\linewidth]{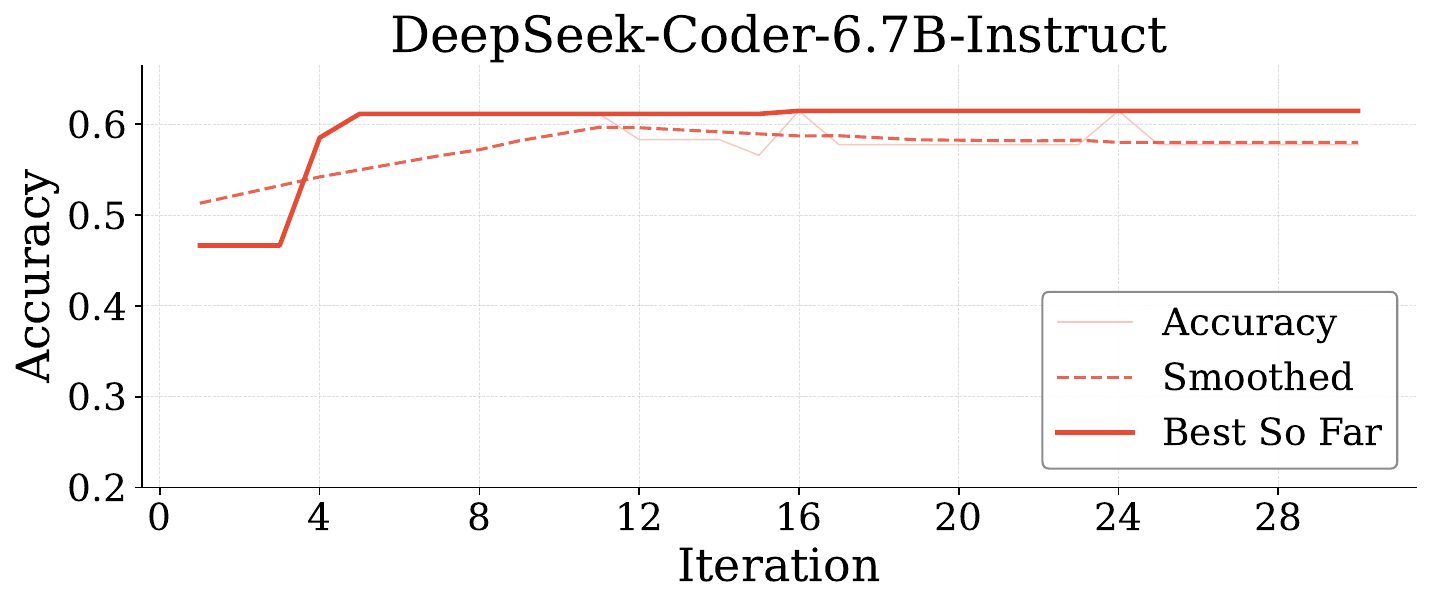}
			\caption{DeepSeek-Coder (ImageNette)}
			\label{fig:acc_deepseek_imagenette}
		\end{subfigure}
		\hfill
		\begin{subfigure}[b]{0.325\textwidth}
			\centering
			\includegraphics[width=\linewidth]{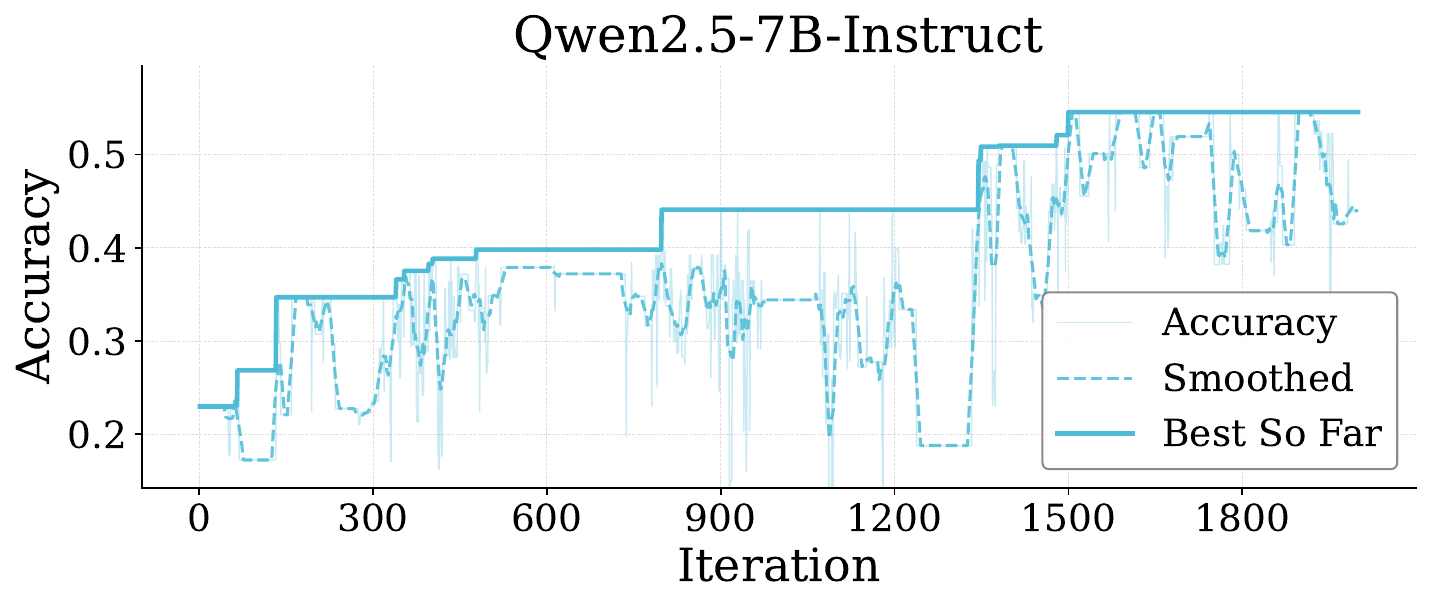}
			\caption{Qwen2.5 (ImageNette)}
			\label{fig:acc_qwen_imagenette}
		\end{subfigure}
		\hfill
		\begin{subfigure}[b]{0.325\textwidth}
			\centering
			\includegraphics[width=\linewidth]{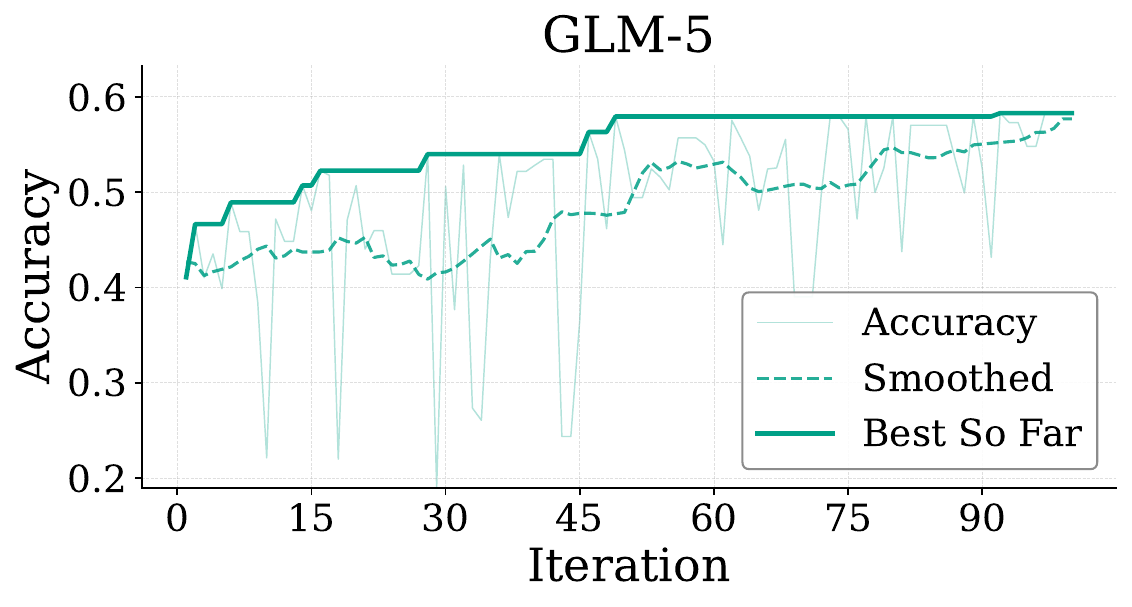}
			\caption{GLM-5 (ImageNette)}
			\label{fig:acc_glm5_imagenette}
		\end{subfigure}
		
		\caption{One-epoch proxy accuracy on \textbf{CIFAR-10} (top row, a--c), \textbf{CIFAR-100} (middle row, d--f), and \textbf{ImageNette} (bottom row, g--i) across all iterations. Light curves show per-iteration accuracy (the accuracy of iterations with errors fall back to previous value), dashed lines show the smoothed trend (window $w{=}15$), and bold lines show the best-so-far trajectory. All models exhibit clear upward trends. For DeepSeek-Coder on ImageNette, only the first 30 iterations are plotted because all subsequent iterations resulted in errors.}
		\label{fig:accuracy_combined}
	\end{figure*}
	
	Neural Architecture Search (NAS) has emerged as a powerful paradigm for automating the design of deep neural networks, achieving competitive or superior performance to hand-crafted architectures across diverse tasks~\cite{Zoph2017,Real2019,Liu2019DARTS}. However, early conventional NAS methods are notoriously resource-intensive: early reinforcement learning approaches required up to 22,400 GPU-days~\cite{Zoph2017}, evolutionary algorithms demanded thousands of GPU-days~\cite{Real2019}, and even efficient differentiable relaxations~\cite{Liu2019DARTS} depend on heavy supernet training. This severe computational cost has motivated data-efficient and training-free approaches that use proxy metrics to rank architectures without full training~\cite{Mellor2021,Li2024ZeroShot}, yet these methods still operate within constrained, predefined search spaces (e.g., cell-based structures).
	
	Recent advances in large language models (LLMs) have opened a fundamentally different avenue: using LLMs as architecture generators that produce executable neural network code directly~\cite{ABrain.NNGPT,ABrain.Architect,ABrain.Prompt}. Unlike traditional NAS, LLM-based approaches operate in the \emph{unconstrained open code space} of object-oriented programs rather than fixed cell-based encodings, enabling more flexible and expressive architectural invention. This efficiency-oriented perspective is central to our work:
by operating on a single consumer-grade GPU with frozen
${\leq}7$B LLMs and no fine-tuning, our pipeline makes
architecture search accessible in resource-constrained
environments and implicitly favors compact models suited
to deployment on hardware-limited devices. Prior work within the NNGPT framework~\cite{ABrain.NNGPT} and related efforts~\cite{ABrain.NN-Captioning_2025,ABrain.NNGPT-Fractal,ABrain.CV_Channel} have demonstrated that LLMs can generate functional vision models, but predominantly through single-shot or few-shot generation without iterative self-improvement.
	
	A critical limitation of single-shot LLM generation is that it treats architecture design as a one-pass prediction, discarding the evaluation signal entirely. In contrast, human engineers iterate: they build, test, analyze failures, and refine their designs based on accumulated experience. This observation motivates our central question: \emph{Can a small LLM iteratively improve neural network architectures by learning from a structured history of its own attempts?}
	
	We propose a closed-loop pipeline comprising three components: (1)~a \textbf{Code Generator} that prompts an instruction-tuned LLM to produce PyTorch model implementations, (2)~an \textbf{Evaluator} that trains each generated model for a single epoch on CIFAR-10, CIFAR-100~\cite{Krizhevsky2009}, or ImageNette~\cite{imagenette} as a fast proxy for architecture quality, and (3)~a \textbf{Prompt Improver} that analyzes evaluation results alongside a sliding window of recent iteration history---a mechanism we term \emph{historical feedback memory}---to produce targeted improvement suggestions for the next iteration. This design draws on the Markov property: the improvement decision at each step depends on the current best architecture and a bounded window of recent transitions, rather than the full trajectory.
	
	Our main contributions are:
	\begin{itemize}
		\item A closed-loop, iterative NAS pipeline driven by LLMs that progressively discovers better architectures through code generation, evaluation, and prompt refinement.
		\item A \emph{historical feedback memory} mechanism that maintains a sliding window of past improvement attempts, enabling the LLM to mitigate repeating failed strategies and build on successful ones.
		\item An empirical demonstration across three LLMs of different origins and specializations showing that the pipeline consistently improves architecture quality---from 28.2\% to 69.2\% (DeepSeek-Coder-6.7B-Instruct), 50.0\% to 71.5\% (Qwen2.5-7B), and 43.2\% to 62.0\% (GLM-5) on CIFAR-10---requiring only $\sim$18 GPU hours on a single consumer-grade 24GB GPU, establishing a low-budget approach to NAS.
	\end{itemize}
	
	\section{Related Work}
	\label{sec:Related}
	
	\paragraph{Neural Architecture Search.}
	NAS methods aim to automate the design of neural network architectures. Early approaches employed reinforcement learning~\cite{Zoph2017} and evolutionary algorithms~\cite{Real2019} to explore architecture search spaces, but required thousands of GPU hours per search. Parameter-sharing methods such as ENAS~\cite{Pham2018} and differentiable approaches like DARTS~\cite{Liu2019DARTS} substantially reduced search cost by amortizing evaluation across architectures. More recently, \emph{zero-shot} or \emph{training-free} NAS methods~\cite{Mellor2021,Li2024ZeroShot} bypass training entirely by scoring architectures at initialization using proxy metrics such as activation overlap~\cite{Mellor2021} or gradient-based indicators. While effective within their respective search spaces, these methods remain constrained to predefined, discrete architecture parameterizations (e.g., cell-based structures) and cannot generate truly novel architectural patterns.
	
	\paragraph{LLMs for Code Generation and AutoML.}
	Large language models have demonstrated strong capabilities in code generation~\cite{Chen2021Codex}, and their application to machine learning automation is an active research area. The NNGPT framework~\cite{ABrain.NNGPT} uses LLMs as neural network generators, showing that language models can produce functional vision models from textual prompts. Subsequent work explored fractal-inspired architectures~\cite{ABrain.NNGPT-Fractal}, few-shot prompting strategies~\cite{ABrain.Prompt}, novel architecture creativity~\cite{ABrain.Architect}, and non-standard channel priors~\cite{ABrain.CV_Channel}. LLMs have also been applied to hyperparameter optimization~\cite{ABrain.HPGPT}, data transformation design~\cite{ABrain.Transform}, and collaborative vision-language pipelines~\cite{Rupani2025llm,Gado2025llm,ABrain.NN-RAG}. These efforts are supported by curated datasets of neural network architectures~\cite{ABrain.NN-Dataset,ABrain.LEMUR2} and deployment pipelines~\cite{ABrain.NN-Lite}. However, most existing approaches treat LLM-based architecture generation as a single-step process, without leveraging iterative refinement from evaluation feedback. Our work extends this line by introducing a closed-loop pipeline with historical memory that enables progressive architecture improvement.
	
	\paragraph{Iterative LLM Optimization.}
	Recent work has explored using LLMs as iterative optimizers that progressively 
	improve solutions by learning from historical attempts. OPRO~\cite{Yang2023OPRO} 
	demonstrated that LLMs can optimize text prompts and discrete solutions by 
	conditioning on sorted history of past candidates and scores. FunSearch~\cite{Romera2023FunSearch} 
	extended this to program synthesis for mathematical discovery using island-based 
	evolutionary search. EvoPrompting~\cite{Chen2023EvoPrompting} applied evolutionary 
	prompting with LLMs as mutation operators for code-level neural architecture 
	search. ReEvo~\cite{Ye2024ReEvo} introduced dual-layer reflective memory for 
	LLM-driven heuristic algorithm design. More directly related to NAS, 
	LLMO~\cite{Zhong2024LLMO} used LLMs as combinatorial optimizers within the 
	predefined cell-based search space of NAS-Bench-201~\cite{dong2020nas}, replacing traditional 
	meta-heuristics. However, existing methods face key limitations: reliance on 
	global elite histories that discard crucial failure trajectories~\cite{Yang2023OPRO,Romera2023FunSearch}, 
	the computational overhead of maintaining evolutionary populations~\cite{Chen2023EvoPrompting}, 
	operation in constrained discrete search spaces~\cite{Zhong2024LLMO}, or 
	targeting domains without structural code constraints~\cite{Ye2024ReEvo}. Our work introduces 
	bounded Markovian memory specifically designed for iterative neural architecture 
	code generation in open search spaces, with explicit failure modeling enabling 
	effectiveness on small frozen LLMs.

	\paragraph{Positioning relative to prior work.}
	\begin{table*}[t]
	\centering
	\fontsize{7.5}{9}\selectfont
	\begin{tabular}{l c c c c c}
	\toprule
	\textbf{Method} & \textbf{NAS target} & \makecell{\textbf{Unconstrained}\\\textbf{search space}} & \makecell{\textbf{Automated failure}\\\textbf{diagnosis}} & \makecell{\textbf{Single-GPU}\\\textbf{($\leq$24\,GB)}} & \makecell{\textbf{No LLM}\\\textbf{fine-tuning}} \\
	\midrule
	OPRO~\cite{Yang2023OPRO} & $\times$ & $\times$ & $\times$ & $\times$ & $\checkmark$ \\
	FunSearch~\cite{Romera2023FunSearch} & $\times$ & $\checkmark$ & $\times$ & $\times$ & $\checkmark$ \\
	EvoPrompting~\cite{Chen2023EvoPrompting} & $\checkmark$ & $\checkmark$ & $\times$ & $\checkmark$ & $\times$ \\
	ReEvo~\cite{Ye2024ReEvo} & $\times$ & $\checkmark$ & $\checkmark$ & $\times$ & $\checkmark$ \\
	LLMO~\cite{Zhong2024LLMO} & $\checkmark$ & $\times$ & $\times$ & $\times$ & $\checkmark$ \\
	\midrule
	\textbf{Ours} & $\checkmark$ & $\checkmark$ & $\checkmark$ & $\checkmark$ & $\checkmark$ \\
	\bottomrule
	\end{tabular}
	\caption{Systematic comparison with most similar LLM-based NAS and optimization methods.
	NAS target: method aims at neural architecture search;
	Unconstrained search space: search over executable code rather than predefined cell encodings or fixed discrete spaces;
	Automated failure diagnosis: structured modeling of code execution failures (e.g., diagnostic triples) as first-class feedback signals;
	Single-GPU ($\leq$24\,GB): demonstrated effective with $\leq$7B LLMs on a single consumer GPU;
	No LLM fine-tuning: operates with frozen pretrained LLMs without any parameter updates.
	Our method uniquely combines all five dimensions.}
	\label{table:comparison}
	\end{table*}

	Table~\ref{table:comparison} positions our method relative to the five most similar LLM-based optimization approaches. We are the only method that simultaneously satisfies all five properties: (1)~targeting neural architecture search, (2)~operating in an unconstrained code search space rather than predefined cells, (3)~automated failure diagnosis via structured diagnostic triples, (4)~demonstrated effectiveness on a single consumer GPU with $\leq$7B LLMs, and (5)~requiring no LLM fine-tuning whatsoever. Notably, while EvoPrompting also performs code-level NAS on small models, it requires soft prompt-tuning of the LLM, whereas our pipeline operates on fully frozen instruction-tuned models. LLMO also applies LLMs to NAS without training, but operates in predefined discrete search spaces that fundamentally limit architectural expressiveness.

	\section{Methodology}
	\label{sec:Methodology}
	
	Inspired by recent advancements in the application of LLMs across various domains~\cite{ABrain.HPGPT,Gado2025llm,Rupani2025llm,ABrain.NN-RAG} and prior architectural synthesis experiments within the NNGPT framework~\cite{ABrain.HPGPT,ABrain.NN-Captioning_2025,ABrain.Prompt,ABrain.NNGPT-Fractal,ABrain.Transform,ABrain.Architect,ABrain.CV_Channel}, and leveraging the existing LEMUR dataset of a broad range of high-capacity and edge-optimized models~\cite{ABrain.NN-Dataset,ABrain.LEMUR2,ABrain.NN-Lite}, we developed an iterative NAS pipeline. The pipeline consists of three core modules operating in a closed loop, as illustrated in Figure~\ref{fig:pipeline}.
	
	\subsection{Pipeline Overview}
	
	At each iteration $t$, the pipeline executes the following cycle (see Algorithm~\ref{alg:pipeline}): (1)~the Code Generator produces a candidate architecture $\mathcal{A}_t$ as executable PyTorch code; (2)~the validated generated code is trained and evaluated by the Evaluator to obtain a proxy accuracy $a_t$; (3)~the Prompt Improver analyzes the result in the context of the current best architecture and recent iteration history, producing improvement suggestions $s_t$ for the next iteration. The best-performing architecture $\mathcal{A}^* = \arg\max_{i \leq t} \mathcal{A}_i$ is maintained as the reference implementation throughout the search.
    
	To support reproducibility, the full pipeline implementation is made available at an anonymized repository: \url{https://anonymous.4open.science/r/Iterative-LLM-Based-NAS-with-Feedback-Memory-E7D6/README.md}.\footnote{The repository includes all prompt templates, training scripts, evaluation harnesses, and per-iteration accuracy logs for all reported experiments.}

	\subsection{Code Generator}
	
	The Code Generator uses a pre-trained, instruction-tuned LLM to produce a complete PyTorch model class implementing \texttt{nn.Module}. The LLM receives a fixed prompt template comprising: (i)~a role description positioning the model as a ``visionary deep learning architect,'' (ii)~the task specification (CIFAR-10, CIFAR-100 or ImageNette image classification without pre-trained weights), (iii)~the current best implementation $\mathcal{A}^*$ as reference code, and (iv)~the improvement suggestions $s_{t-1}$ from the previous iteration. The generated code must define a \texttt{Net(nn.Module)} class with standard \texttt{\_\_init\_\_} and \texttt{forward} methods. Generation uses temperature $\tau = 0.7$ and nucleus sampling ($p = 0.9$) to balance architectural diversity with code coherence.

For large generalist models with sufficiently large context
windows (in our experiments, GLM-5), we additionally
evaluate an extended prompt format that supplements the
standard template with the implementations of the top-$K$
best-performing architectures discovered so far, alongside
the executable code and evaluation outcomes of the
previous $K$ iterations. This variant provides richer
positive exemplars at the cost of increased prompt length,
and is applicable only when the model's context window
can accommodate the additional content without truncation.
All other pipeline components remain identical.
	
	\subsection{Evaluator}
	
	Generated code undergoes a two-stage evaluation process.
	
	\textbf{Quick validation.} The model is instantiated and a forward pass is performed with a dummy input tensor of shape corresponding to the dataset (e.g., $2 \times 3 \times 32 \times 32$ for CIFAR-10). The output shape is verified to match $B \times C$ (for batch size $B$ and $C$ classes). Models failing this check are discarded with an error message forwarded to the Prompt Improver.
	
	\textbf{Proxy training.} Models passing validation are trained for \textbf{one epoch} on the respective dataset (CIFAR-10, CIFAR-100 or ImageNette) using SGD (momentum $0.9$, weight decay $5 \times 10^{-4}$), an initial learning rate of $0.01$ with cosine annealing, and a batch size of $128$. Standard data augmentation is applied: random crop with padding and random horizontal flip. Top-1 test accuracy serves as the proxy metric. This one-epoch accuracy provides a fast, informative signal for ranking architectures~\cite{ru2021speedy}, enabling rapid iteration without the cost of full convergence training.
	
	\subsection{Prompt Improver with Historical Feedback Memory}
	\label{sec:history}
	
	The Prompt Improver is the central component enabling iterative learning. After each evaluation, it receives three inputs: (1)~the best code $\mathcal{A}^*$ and its accuracy $a^*$, (2)~the current iteration's code $\mathcal{A}_t$ and its evaluation outcome (accuracy value or error message), and (3)~a \emph{historical feedback memory} $\mathcal{H}_t$ containing the last $K{=}5$ improvement attempts.
	
	Each history entry $h_i \in \mathcal{H}_t$ is a triple recording: the identified problem found in the previous iteration about the generated code, the suggested improvement, and the resulting outcome (accuracy achieved or error encountered). This sliding-window design draws on the Markov property: the improvement decision at step $t$ depends on the current state (best architecture plus recent history window) rather than the complete trajectory. The bounded window prevents context overflow while providing sufficient signal for the LLM to identify failure patterns.
	
	\paragraph{Markov Property Formalization.}
	Let $\mathcal{A}_t$ denote the architecture generated at step $t$, $a_t$ its 
	evaluation outcome (accuracy or error), and $s_t$ the improvement suggestions. 
	Our historical feedback memory $\mathcal{H}_t^{(K)}$ maintains exactly the most 
	recent $K{=}5$ improvement attempts:
	\begin{equation}
	\mathcal{H}_t^{(K)} = \{(s_{t-K}, a_{t-K}), \ldots, (s_{t-1}, a_{t-1})\}.
	\end{equation}
	The improvement suggestion generation satisfies the \textbf{K-order Markov property}:
	\begin{equation}
	P(s_t \mid \mathcal{A}^*, \mathcal{H}_t) = P(s_t \mid \mathcal{A}^*, \mathcal{H}_t^{(K)}),
	\end{equation}
	meaning the next suggestion depends only on the current best architecture and 
	the bounded recent history, not the complete trajectory. This design ensures 
	\textbf{constant context size} and avoids the context overflow problem faced 
	by unbounded history approaches like OPRO~\cite{Yang2023OPRO}.
	
	Each history entry $(s_i, a_i)$ is structured as a \textbf{diagnostic triple}:
	\begin{equation}
	s_i = (\text{problem}_i, \text{suggestion}_i, \text{outcome}_i),
	\end{equation}
	where $\text{problem}_i$ identifies the architectural deficiency, 
	$\text{suggestion}_i$ proposes concrete code modifications, and $\text{outcome}_i$ 
	records whether the suggestion succeeded (accuracy gain) or failed (error type). 
	This structured format enables the LLM to learn causal patterns between design 
	decisions and outcomes, unlike scalar-score histories in prior work~\cite{Yang2023OPRO,Romera2023FunSearch}.
	
	The Prompt Improver produces a structured response containing: (a)~a \textbf{reason} diagnosing the current result, (b)~an \textbf{inspiration} drawing on cross-disciplinary insights, and (c)~concrete \textbf{improvement suggestions} for the next iteration. These suggestions, together with the best code and the code of current iteration, form the input to the Code Generator in the subsequent iteration, closing the feedback loop.

	\begin{algorithm}[t]
		\caption{Iterative LLM-Driven NAS Pipeline}
		\label{alg:pipeline}
		\KwIn{LLM $\mathcal{L}$, max iterations $T$, history window $K$}
		$\mathcal{A}^* \leftarrow \emptyset$; $a^* \leftarrow 0$; $\mathcal{H} \leftarrow \emptyset$; $s_0 \leftarrow \emptyset$\;
		\For{$t \leftarrow 1$ \KwTo $T$}{
			$\mathcal{A}_t \leftarrow \text{Generate}(\mathcal{L}, \mathcal{A}^*, s_{t-1})$\;
			\eIf{$\text{Validate}(\mathcal{A}_t)$ succeeds}{
				$a_t \leftarrow \text{TrainAndEvaluate}(\mathcal{A}_t)$\;
			}{
				$a_t \leftarrow$ error message\;
			}
			\If{$a_t > a^*$}{
				$\mathcal{A}^* \leftarrow \mathcal{A}_t$; $a^* \leftarrow a_t$\;
			}
			Append $(s_{t-1}, a_t)$ to $\mathcal{H}$; keep last $K$ entries\;
			$s_t \leftarrow \text{Improve}(\mathcal{L}, \mathcal{A}^*, \mathcal{A}_t, a_t, \mathcal{H})$\;
		}
		\Return $\mathcal{A}^*$, $a^*$
	\end{algorithm}

	\section{Experiments}
	\label{sec:experiments}
	
	\subsection{Experimental Setup}
	
	\paragraph{Dataset.} We evaluate our approach on three datasets: CIFAR-10~\cite{Krizhevsky2009}, CIFAR-100~\cite{Krizhevsky2009}, and ImageNette~\cite{imagenette}. CIFAR-100 shares the same $32{\times}32$ image dimensions as CIFAR-10 but increases the classification difficulty from 10 to 100 fine-grained classes. Standard augmentation (random crop with padding, random horizontal flip, normalization) is applied during training, inspired by~\cite{Aboudeshish2025augmentation}.
	
	\paragraph{Language models.} To evaluate the pipeline across model specializations, we run separate experiments with three LLMs: code-specialized DeepSeek-Coder-6.7B-Instruct, small generalist Qwen2.5-7B-Instruct, and large generalist GLM-5. All models use the same generation parameters: temperature $0.7$, nucleus sampling $p = 0.9$, maximum 2,048 new tokens. A fixed seed with incremental call counter ensures reproducible but diverse generation across iterations. DeepSeek-Coder-6.7B and Qwen2.5-7B are each run for 2000 iterations; GLM-5 is only run for 100 iterations for economic reasons.
	
	\paragraph{Training protocol.} Each candidate architecture is trained for one epoch with SGD (momentum $0.9$, weight decay $5 \times 10^{-4}$), learning rate $0.01$ with cosine annealing, and batch size $128$. A fixed random seed ($43$) and deterministic CUDA operations ensure reproducibility. Training runs in isolated subprocesses with a 30-minute timeout. Fine-tuning of LLMs and training of computer vision models are performed on a NVIDIA GeForce RTX 4090 24G GPU.
	
	\paragraph{Baseline.} We compare against \textbf{single-shot generation}: the accuracy of the first successfully generated and evaluated architecture (iteration~1), representing what the LLM produces from its pre-trained knowledge alone without any feedback signal.
	
	\subsection{Evaluation Metrics}
	\label{sec:metrics}
	
	To evaluate our approach rigorously, we adopt true validation accuracy after the first training epoch as our primary architecture quality metric~\cite{egele2024unreasonable}, rather than relying on zero-shot NAS proxies. Although zero-shot proxies are correlated with fully trained accuracy~\cite{Mellor2021, chen2021tenas, lin2021zen}, they remain indirect indicators of performance. Even the strongest reported Spearman rank-correlation coefficients ($\rho \approx 0.5$--$0.82$) on standard benchmarks such as NAS-Bench-101~\cite{ying2019nas} and NAS-Bench-201~\cite{dong2020nas} correspond to a coefficient of determination of at most $R^2 \approx 0.67$, leaving substantial variance unexplained~\cite{abdelfattah2021zerocost, white2023neural, yu2020evaluating, yin2022bmnas}. By using first-epoch validation accuracy instead, we aim to demonstrate more directly that our algorithm can reliably influence and accelerate early-stage performance trajectories of neural networks. 
	
	To characterize search dynamics, we compute: Spearman rank correlation $\rho$ and Kendall $\tau$ between iteration index and accuracy. All trends are evaluated for statistical significance.

	\subsection{Main Results}
	
	Table~\ref{table:summary_combined} compares the key results across the LLMs. The models exhibit statistically upward trends in architecture quality across all three datasets (Spearman $\rho$ up to $0.75$, $p \approx 0$), confirming that the iterative pipeline with historical feedback memory consistently drives improvement.
	
	\paragraph{DeepSeek-Coder-6.7B-Instruct.} Among the three models, the code-specialized DeepSeek-Coder stands out for its remarkable reliability on CIFAR-class datasets: it completed 1519 of 2000 iterations successfully on CIFAR-10 (76.0\%) and an even higher 1902 of 2000 on CIFAR-100 (95.1\%), suggesting that the structured feedback loop becomes more effective as the model accumulates experience with similar low-resolution inputs. This reliability translates into strong improvement trajectories---on CIFAR-10, one-epoch proxy accuracy climbed from a 28.2\% baseline to 69.2\% ($\rho = 0.754$), the largest absolute gain observed across all model--dataset combinations. On CIFAR-100, it progressed from 5.0\% to 29.2\% ($\rho = 0.205$, $p \approx 0$), with a notably higher success rate than on CIFAR-10 despite the increased class count. However, this consistency collapsed entirely on ImageNette: the higher input resolution ($160{\times}160$) triggered persistent context retention failures, leaving only 13 successful evaluations out of 2000 (0.7\%). Even so, the few successful architectures reached 61.5\% accuracy from an initial 46.7\%, indicating that the model's code quality remained high when generation succeeded.

	\paragraph{Qwen2.5-7B-Instruct.} At first glance, Qwen2.5-7B appears to be the weakest performer: its CIFAR-10 success rate of just 18.8\% (376 of 2000 iterations) is by far the lowest among the three models. Yet this generalist model ultimately achieved the highest peak accuracy on CIFAR-10---71.5\%, surpassing both DeepSeek-Coder (69.2\%) and GLM-5 (62.0\%). This apparent paradox reflects a distinctive exploration strategy: over thousands of iterations, Qwen2.5 increasingly generates ambitious architectures, many of which fail but occasionally yield superior designs. The Spearman correlation remains significant ($\rho = 0.561$, $p \approx 0$), confirming genuine improvement despite the noise. A similar pattern emerges on CIFAR-100, where it maintained a moderate 53.2\% success rate (1064 of 2000) and improved from 27.0\% to 29.6\%, though the weaker correlation ($\rho = 0.037$) suggests that extended error loops on this harder task limited effective search diversity. On ImageNette, the success rate improved to 32.1\% (642 evaluations), and proxy accuracy rose substantially from 23.0\% to 54.6\% ($\rho = 0.663$)---the strongest correlation among all model--dataset pairs for this benchmark.

	\paragraph{GLM-5.} Running for only 100 iterations, the large generalist GLM-5 from ZhipuAI provides evidence that extensive iteration counts are not strictly necessary for large language models. It achieved the highest success rates across all three datasets---91.0\% on CIFAR-10, 85.0\% on CIFAR-100, and 81.0\% on ImageNette---reflecting a conservative generation strategy that favors reliable, well-structured architectures. On CIFAR-10, proxy accuracy rose from 43.2\% to 62.0\% ($\rho = 0.422$), with improvement continuing throughout the search and the peak reached at the 91st successful evaluation. The same convergence pattern appeared on CIFAR-100 (19.1\% $\to$ 24.8\%, $\rho = 0.431$) and ImageNette (41.1\% $\to$ 58.3\%, $\rho = 0.631$). While GLM-5's peak accuracies are below those of models given 20$\times$ more iterations, its high success rate and consistent improvement make it the most reliable search agent in our evaluation.

	Figure~\ref{fig:accuracy_combined} shows the accuracy trajectories for all models across all three datasets. All curves display consistent upward trends: DeepSeek-Coder achieves the largest absolute improvement on CIFAR-10, Qwen2.5 reaches the highest peak accuracy, and GLM-5 exhibits a robust upward trend across all three datasets. The pipeline generalizes well across datasets of varying difficulty, confirming its broad applicability.
	
	\begin{table*}[t]
		\centering
		\fontsize{7.5}{8.5}\selectfont
		\setlength{\tabcolsep}{4pt}
		
		\begin{minipage}[t]{0.32\textwidth}
			\centering
			\centerline{\textbf{(a) CIFAR-10}}
			\vspace{0.1cm}
			\begin{tabular}{lrrr}
				\toprule
				\textbf{Metric} & \textbf{DeepS.} & \textbf{Qwen} & \textbf{GLM} \\
				\midrule
				Total iters & 2000 & 2000 & 100 \\
				Succ.\ evals & 1519 & 376 & 91 \\
				Succ.\ rate & 76.0\% & 18.8\% & 91.0\% \\
				\midrule
				1st acc. & 28.2\% & 50.0\% & 43.2\% \\
				Best acc. & 69.2\% & 71.5\% & 62.0\% \\
				Improve. & +41.0\% & +21.5\% & +18.7\% \\
				\midrule
				$\rho$ & $0.754$ & $0.561$ & $0.422$ \\
				$\tau$ & $0.551$ & $0.407$ & $0.300$ \\
				\bottomrule
			\end{tabular}
		\end{minipage}
		\hfill
		\begin{minipage}[t]{0.32\textwidth}
			\centering
			\centerline{\textbf{(b) CIFAR-100}}
			\vspace{0.1cm}
			\begin{tabular}{lrrr}
				\toprule
				\textbf{Metric} & \textbf{DeepS.} & \textbf{Qwen} & \textbf{GLM} \\
				\midrule
				Total iters & 2000 & 2000 & 100 \\
				Succ.\ evals & 1902 & 1064 & 85 \\
				Succ.\ rate & 95.1\% & 53.2\% & 85.0\% \\
				\midrule
				1st acc. & 5.0\% & 27.0\% & 19.1\% \\
				Best acc. & 29.2\% & 29.6\% & 24.8\% \\
				Improve. & +24.1\% & +2.7\% & +5.7\% \\
				\midrule
				$\rho$ & $0.205$ & $0.037$ & $0.431$ \\
				$\tau$ & $0.142$ & $0.031$ & $0.311$ \\
				\bottomrule
			\end{tabular}
		\end{minipage}
		\hfill
		\begin{minipage}[t]{0.32\textwidth}
			\centering
			\centerline{\textbf{(c) ImageNette}}
			\vspace{0.1cm}
			\begin{tabular}{lrrr}
				\toprule
				\textbf{Metric} & \textbf{DeepS.} & \textbf{Qwen} & \textbf{GLM} \\
				\midrule
				Total iters & 2000 & 2000 & 100 \\
				Succ.\ evals & 13 & 642 & 81 \\
				Succ.\ rate & 0.7\% & 32.1\% & 81.0\% \\
				\midrule
				1st acc. & 46.7\% & 23.0\% & 41.1\% \\
				Best acc. & 61.5\% & 54.6\% & 58.3\% \\
				Improve. & +14.8\% & +31.6\% & +17.2\% \\
				\midrule
				$\rho$ & $-0.084$ & $0.663$ & $0.631$ \\
				$\tau$ & $-0.069$ & $0.444$ & $0.452$ \\
				\bottomrule
			\end{tabular}
		\end{minipage}
		\caption{Comparison of the iterative NAS pipeline across three LLMs on CIFAR-10, CIFAR-100, and ImageNette datasets. The models exhibit statistically significant improvement over single-shot generation. On ImageNette, correlation metrics for DeepSeek are not meaningful due to insufficient successful evaluations. On CIFAR-100, Qwen2.5 shows weaker correlation due to extended error loops reducing effective search diversity.}
		\label{table:summary_combined}
	\end{table*}

	\section{Ablation Study}
	\label{sec:ablation}
    
	\begin{figure*}[t]
		\centering
		\begin{subfigure}[b]{0.32\textwidth}
			\centering
			\includegraphics[width=\linewidth]{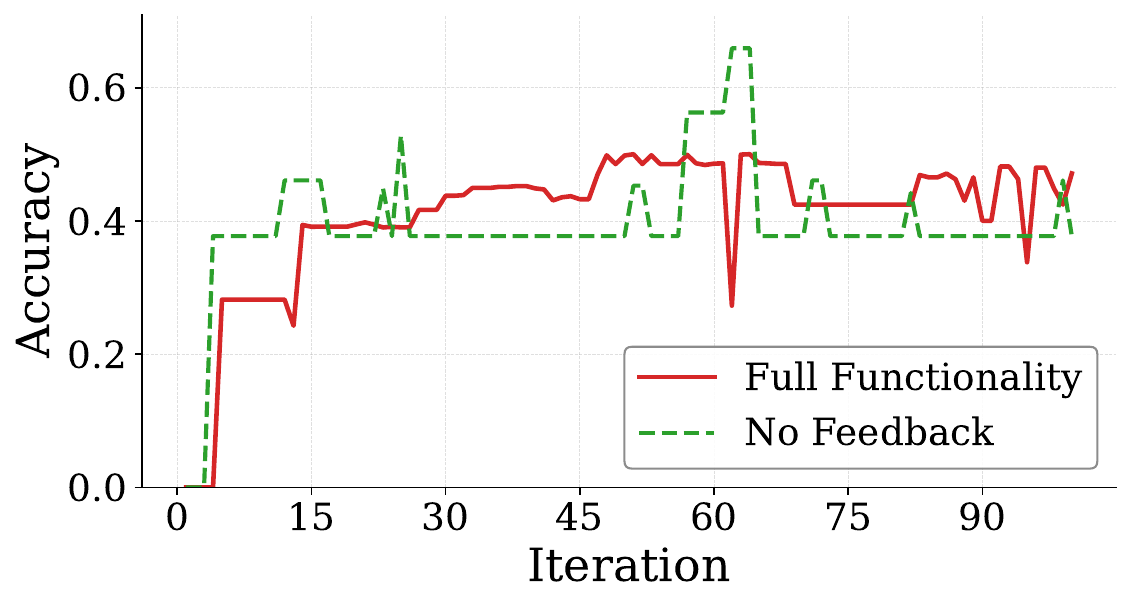}
			\caption{CIFAR-10}
			\label{fig:ablation_deepseek_cifar}
		\end{subfigure}
		\hfill
		\begin{subfigure}[b]{0.32\textwidth}
			\centering
			\includegraphics[width=\linewidth]{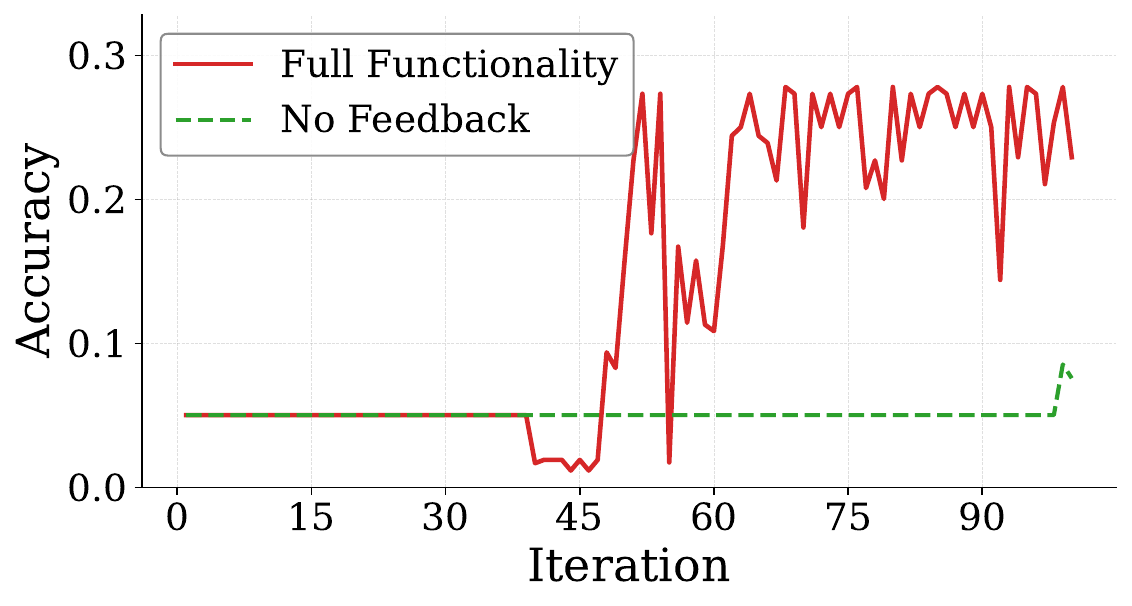}
			\caption{CIFAR-100}
			\label{fig:ablation_deepseek_cifar100}
		\end{subfigure}
		\hfill
		\begin{subfigure}[b]{0.32\textwidth}
			\centering
			\includegraphics[width=\linewidth]{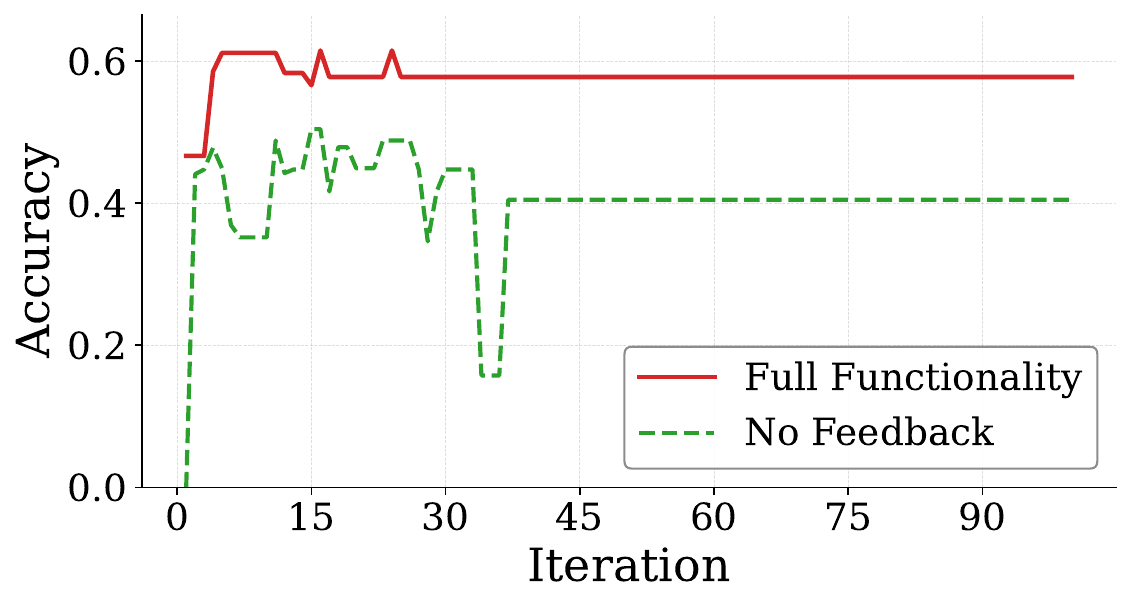}
			\caption{ImageNette}
			\label{fig:ablation_deepseek_imagenette}
		\end{subfigure}
		\caption{Ablation study of DeepSeek-Coder-6.7B-Instruct on CIFAR-10, CIFAR-100, and ImageNette datasets. The results highlight the effectiveness of the complete iterative loop with historical feedback memory compared to its ablated variants.}
		\label{fig:ablation_deepseek}
	\end{figure*}

	Additionally, we conducted ablation experiments (Figure~\ref{fig:ablation_deepseek}) involving DeepSeek-Coder-6.7B on CIFAR-10, CIFAR-100, and ImageNette to evaluate how removing historical feedback or reference architectures impacts the search trajectory. The results show that without the historical feedback memory or the reference architecture, the search process stagnates or degrades significantly and fails to surpass the single-shot baseline. The anomalous accuracy spike ($\sim$66\%) in the CIFAR-10 \emph{No Feedback} trajectory is an incidental discovery that is swiftly lost without history retention. This underscores that explicitly modeling causality from past code execution failures is critical for iterative architectural improvement.
	
	\section{Discussion}
	\label{sec:discussion}
	
	\paragraph{Effectiveness of iterative refinement.}
	All models exhibit statistically significant positive correlations between evaluation index and accuracy ($\rho$ up to $0.75$, Kendall $\tau$ up to $0.55$, all $p \approx 0$), confirming that historical feedback memory enables systematic architectural improvement across different LLM origins and specializations. Access to the last $K{=}5$ improvement attempts provides the LLM with sufficient context to identify recurring failure patterns and avoid previously unsuccessful strategies. All accuracy curves display characteristic increasing behavior---rapid early gains followed by gradual flattening---mirroring patterns in traditional optimization and suggesting that the LLM-driven search exhibits meaningful optimization dynamics rather than random exploration.
	
	\paragraph{Effect of model specialization and iteration scale.}
	The three models reveal distinct behavior along two axes. DeepSeek-Coder-6.7B achieves the largest absolute improvement on CIFAR-10 (+41.0 pp from single-shot to peak) with a high and stable success rate (76.0\%) over 2000 iterations, suggesting that code specialization yields both reliable generation and systematic progression in specific situations. Qwen2.5-7B achieves the highest peak accuracy (71.5\%) at the cost of a dramatically lower success rate (18.8\%) over 2000 iterations---its Markovian exploration increasingly targets complex, ambitious architectures that more often fail validation but, when successful, surpass DeepSeek-Coder's best. GLM-5 demonstrates consistent improvement ($+18.7$ pp from 43.2\% to 62.0\%) with the highest success rate (91.0\%) within only 100 iterations, making it the most reliable search agent of the three.
	
	\paragraph{Comparison with iterative LLM optimization methods.}
	Our approach shares conceptual similarities with recent work on using LLMs as 
	iterative optimizers~\cite{Yang2023OPRO,Romera2023FunSearch,Ye2024ReEvo}, but 
	differs in two critical dimensions that enable its effectiveness on neural 
	architecture search:
	
	\textbf{Structured failure modeling.} Our diagnostic triples explicitly encode 
	code execution failures as first-class entries in the history, enabling 
	the LLM to learn from structural mistakes. This is fundamentally different from 
	prior work where low-performing solutions are simply discarded from the prompt~\cite{Yang2023OPRO,Romera2023FunSearch} 
	or managed implicitly through population selection~\cite{Chen2023EvoPrompting}. 
	Failure rates of 5--99\% across our runs highlight that code generation for neural architectures is inherently 
	error-prone; ignoring these failures would discard critical learning signals.
	
	\textbf{Hardware-aware and low-budget search.} Unlike early traditional NAS methods that required thousands of GPU days~\cite{Zoph2017,Real2019}, our entire 2000-iteration pipeline completes in approximately 18 GPU hours on a single consumer-grade RTX 4090 GPU. Our results demonstrate that $\leq$7B frozen LLMs of diverse origins can effectively perform NAS in severely resource-constrained environments, whereas prior iterative LLM optimization work primarily relies on massive proprietary models (e.g., GPT-4 and PaLM 2-L in OPRO~\cite{Yang2023OPRO}). Furthermore, by explicitly evaluating in a shared environment where the LLM occupies most of the VRAM, the search implicitly discovers hardware-efficient models tailored to limited memory capacities. We hypothesize this is partly enabled 
	by our dual-LLM specialization: the Code Generator focuses solely on code synthesis 
	given structured suggestions, while the Prompt Improver handles diagnostic reasoning. 
	This task decomposition reduces per-call cognitive load compared to end-to-end 
	generation~\cite{Chen2023EvoPrompting}.

	\paragraph{One-epoch proxy as architecture ranking signal.}
	Using one-epoch accuracy as a ranking signal offers a pragmatic trade-off between evaluation cost and informativeness. While the absolute accuracy values are well below state-of-the-art results with full training, prior work on training-free NAS~\cite{Mellor2021,Li2024ZeroShot} has established that cheap proxy metrics can effectively rank architectures. Our one-epoch proxy serves a similar role: it distinguishes between architectures of varying quality at minimal cost, enabling the feedback loop to guide the LLM toward better designs. Retraining the best-discovered architectures with full training schedules could verify whether the one-epoch ranking transfers to converged performance.
	
	\paragraph{Limitations.}
	Our study has several limitations. First, the code generation success rates vary substantially (0.7--95.1\%), and the causes of model-specific failure rates at scale (e.g., Qwen2.5's 18.8\% success rate over 2000 iterations on CIFAR-10) merit further investigation into prompt design and constrained decoding strategies. Second, while our evaluation now spans three datasets (CIFAR-10, CIFAR-100, and ImageNette), experiments on larger-scale benchmarks and diverse tasks would better demonstrate versatility.
	
	\paragraph{Future work.}
	Promising directions include: (1)~retraining the top-$K$ discovered architectures with full training schedules to verify their converged performance; (2)~extending the cross-model comparison to additional LLM families and sizes;  (3)~extending to larger datasets (e.g., ImageNet) and tasks beyond image classification; and (4) incorporating evolutionary selection into the historical feedback memory mechanism.

	\section{Conclusion}
	\label{sec:conclusion}
We presented an iterative NAS pipeline that leverages instruction-tuned LLMs to progressively generate and improve neural network architectures through a closed loop of code generation, evaluation, and prompt refinement with historical feedback memory. Unlike prior iterative LLM optimization methods that rely on global elite retention or operate in constrained discrete search spaces, our pipeline operates entirely in an unconstrained open code space over executable PyTorch code, enabling genuinely novel architectural patterns that fixed cell-based encodings cannot express. Evaluated on CIFAR-10, CIFAR-100, and ImageNette with three frozen instruction-tuned LLMs (${\leq}7$B parameters) across up to 2000 iterations, the pipeline significantly improved one-epoch proxy accuracy. On CIFAR-10, performance improved from 28.2\% to a peak of 69.2\% for DeepSeek-Coder-6.7B ($\rho = 0.75$), from 50.0\% to 71.5\% for Qwen2.5-7B ($\rho = 0.56$), and from 43.2\% to 62.0\% for GLM-5 ($\rho = 0.42$). On CIFAR-100, DeepSeek-Coder achieved the largest improvement from 5.0\% to 29.2\%, while GLM-5 showed a strong trend ($\rho = 0.43$) despite only 100 iterations. ImageNette results confirmed these positive trends (e.g., Qwen2.5-7B improved from 23.0\% to 54.6\%, $\rho = 0.66$), demonstrating that iterative feedback with bounded historical memory significantly outperforms single-shot generation across model origins and datasets of varying difficulty.

Four design principles distinguish our approach from prior iterative LLM optimization methods. First, the pipeline operates in an open code space rather than constrained predefined cells, enabling expressive and diverse architectural invention. Second, a bounded $K{=}5$-step Markovian memory keeps context size constant, preventing overflow while retaining sufficient failure patterns for systematic improvement --- a lightweight design that removes any dependency on large context windows or proprietary frontier models. Third, structured diagnostic triples explicitly model code execution failures as first-class entries in the history, recording the identified problem, suggested modification, and resulting outcome; this causal structure enables the LLM to avoid repeating failed strategies, which is critical given failure rates of 5--99\% observed across runs. Fourth, a dual-LLM specialization --- a \emph{Code Generator} focused on architecture synthesis and a \emph{Prompt Improver} focused on diagnostic reasoning --- reduces per-call cognitive load and, since both roles share the same limited VRAM with architecture training, implicitly biases the search toward compact, hardware-efficient models well suited to edge deployment.

A full 2000-iteration search completes in ${\approx}18$ GPU hours on a single consumer-grade RTX~4090, with no LLM fine-tuning required at any stage. Our results establish that even ${\leq}7$B LLMs from diverse AI ecosystems can serve as effective architecture search agents when equipped with structured iterative feedback, offering a lightweight, data-efficient, and low-budget paradigm for reproducible NAS that is fully accessible without cloud infrastructure. We believe this positions iterative LLM-driven NAS as a practical tool for researchers operating under hardware constraints, where the low search cost enables discovery of compact architectures that are candidates for deployment on resource-limited devices.
	
	{
		\small
		\bibliographystyle{ieeenat_fullname}
		\bibliography{bibmain}
	}
	
\end{document}